%% file: main.tex
\documentclass{article}
\usepackage{ijcai24}

\usepackage{times}
\usepackage{soul}
\usepackage{url}
\usepackage[hidelinks]{hyperref}
\usepackage[utf8]{inputenc}
\usepackage[small]{caption}
\usepackage{graphicx}
\usepackage{amsmath}
\usepackage{amsthm}
\usepackage{booktabs}
\usepackage{algorithm}
\usepackage{algorithmic}
\usepackage[switch]{lineno}

\usepackage{graphicx}
\usepackage{enumitem}
\usepackage{booktabs}
\usepackage{subcaption}
\usepackage{listings}
\usepackage{multirow}
\usepackage{tcolorbox}
\tcbuselibrary{raster}
\tcbuselibrary{breakable}

\graphicspath{{figs/}}

\newcommand{\vx}{\mathbf{x}}
\newcommand{\vy}{\mathbf{y}}
\newcommand{\vz}{\mathbf{z}}
\newcommand{\mI}{\mathbf{I}}

\definecolor{promptinsert}{HTML}{bfefff}
\definecolor{codehlcolor}{HTML}{ffec8b}
\definecolor{codehlcolor2}{HTML}{ffbbff}

\newcommand{\codehl}[1]{\sethlcolor{codehlcolor}\hl{#1}}

\lstdefinestyle{mystyle}{
    basicstyle=\fontsize{8}{10}\ttfamily,
    keywordstyle=\color{blue},
    commentstyle=\color{gray},
    stringstyle=\color{black},
    showstringspaces=false,
    breaklines=true,
    breakindent=0pt,
    breakatwhitespace=true,
    escapeinside={(*@}{@*)}
}

\title{Structured Code Representations Enable \\ Data-Efficient Adaptation of Code Language Models}

\author{
Mayank Agarwal$^1$
\and
Yikang Shen$^1$\and
Bailin Wang$^{2}$\and
Yoon Kim$^2$\and
Jie Chen$^1$
\affiliations
$^1$MIT-IBM Watson AI Lab, IBM Research $\enspace$ $^2$MIT\\
\emails
mayank.agarwal@ibm.com
}

\begin{document}

\maketitle

\input{sections/abstract}
\input{sections/introduction}
\input{sections/related}
\input{sections/method}
\input{sections/experiment}

\input{sections/conclusion}

\bibliographystyle{named}
\bibliography{references}

\clearpage
\newpage
\appendix
\input{sections/appendix}

\end{document}

%% file: sections/abstract.tex
\begin{abstract}
Current language models tailored for code tasks often adopt the pre-training-then-fine-tuning paradigm from natural language processing, modeling source code as plain text. 
This approach, however, overlooks the unambiguous structures inherent in programming languages.
In this work, we explore data-efficient adaptation of pre-trained code models by further pre-training and fine-tuning them with program structures.
Specifically, we represent programs as parse trees---also known as concrete syntax trees (CSTs)---and adapt pre-trained models on serialized CSTs.
Although the models that we adapt have been pre-trained only on the surface form of programs, we find that a small amount of continual pre-training and fine-tuning on CSTs without changing the model architecture yields improvements over the baseline approach across various code tasks.
The improvements are found to be particularly significant when there are limited training examples, demonstrating the effectiveness of integrating program structures with plain-text representation even when working with backbone models that have not been pre-trained with structures.
\end{abstract}

%% file: sections/introduction.tex
\section{Introduction}
The dominant paradigm in contemporary natural language processing involves pre-training large-scale Transformers~\cite{vaswani2017attention} on surface-form text and then adapting these pre-trained language models to downstream tasks of interest.
This transfer learning paradigm has been extended to programming languages by treating source code as plain text and then applying existing self-learning objectives (e.g., masked-token recovery, next-token prediction) for pre-training and fine-tuning.
This simple approach has been found to be effective when applied to large-scale code data, as demonstrated by the strong performance of existing code language models~\cite{wang2021codet5,chen2021evaluating,li2023starcoder}.
However,  source code inherits well-defined structures that encode the compositional semantics of code.
Unlike natural languages where obtaining analogous structures is challenging due to the lack of an existing grammar, programming languages are paired with grammars and compilers that can unambiguously and efficiently parse a program.
This work explores how the resulting parse tree----concrete syntax tree (CST)---can be exploited to better adapt pre-trained models for code tasks. 

We focus on CSTs rather than other structures (such as control flow graphs~\cite{Allen1970}) for several reasons.
First, many tasks require decoding a program (such as code translation and generation) and it is unclear how to straightforwardly decode a program from  graph representations of code.
Second, off-the-shelf CST parser generators exist for many languages of interest, and support for additional languages is expected to grow;\footnote{\url{https://tree-sitter.github.io/tree-sitter/}} tooling support contributes to the feasibility and practicality of using CSTs.
Third, a CST can be converted to and from a sequence by an easily defined invertible mapping.
Such a property allows for straightforward reuse of the pre-training architecture without modification (i.e., Transformer), in contrast to graph-based representations which often make use of graph-based architectures~\cite{Allamanis2018,Zuegner2021}.

With the CST associated with each program, our approach simply continues pre-training and fine-tuning the pre-trained code model on serialized CSTs.
We experiment with various pre-training objectives to leverage the extensive structural information of CSTs.
In contrast to prior works which use tree-based architectures that decode a tree from parent nodes to children~\cite{Zhang2016,Dong2016,Rabinovich2017,Yin2017,Chen2018,Yin2018,Sun2019,Sun2020}, 
our method adopts a simpler strategy that simply treats the CST as a serialized sequence.
Such an approach is plug-and-play, allowing the reuse of demonstrably scalable frameworks without re-engineering the model architecture; the incurred effort is mostly data preparation and post-processing.

We find that our approach is particularly effective in low-data scenarios.
For example, fine-tuning with structures by using only 100 examples can outperform not using structures by at least 5 CodeBLEU points and 10 percent more passing programs for the code translation task.
Similarly, in the code generation task, structured fine-tuning with 100 examples can achieve a higher CodeBLEU score than the base model fine-tuned with more than 1.8k examples; and in the code summarization task, the structured model using fewer than 200 examples achieves the same performance as the base model using 1k examples.

Our contributions can be summarized as follows.

\begin{enumerate}[leftmargin=*]
\item We introduce a plug-and-play approach for incorporating program structures in the continual pre-training and fine-tuning of language models.
  In particular, we explore the serialized form of CST, considering its suitability as an output format of code for decoding, beyond merely facilitating representation learning.

\item Our structured approach is applicable to both encoder-decoder and decoder-only models.
  Through empirical experiments with both types of models, we observe consistent and beneficial results from our approach.

\item We demonstrate that the structured approach allows data-efficient adaptation of pre-trained models across a variety of tasks, including code translation, generation, and summarization.
\end{enumerate}

%% file: sections/related.tex
\section{Related Work}

\begin{figure*}[t]
  \centering
  \includegraphics[width=0.7\textwidth]{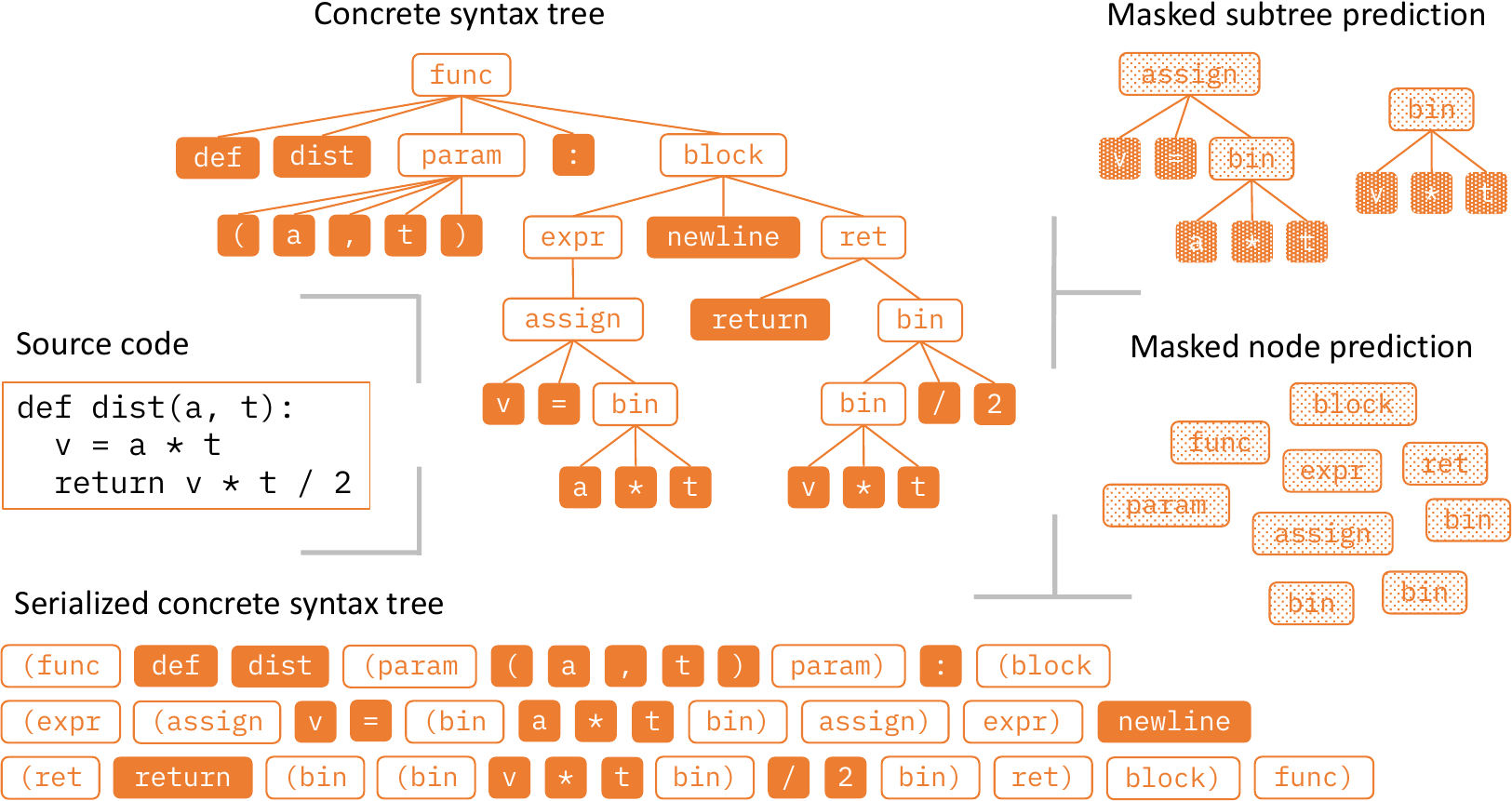}
  \caption{An example Python program along with its CST (simplified for illustration) in the tree and serialized forms, respectively. Also illustrated are the masked subtree prediction and masked node prediction training objectives for adapting pre-trained models to code structures.}
  \label{fig:cst-obj}
\end{figure*}

\paragraph{Machine Learning for Code} 
While early approaches applied statistical learning methods to code tasks \cite{Nguyen2013LexicalSM,movshovitz2013natural,raychev2014code,allamanis2014learning}, with the advent of deep learning, deep neural networks gradually replaced statistical approaches \cite{allamanis2016convolutional,mou2016convolutional,gu2016deep,iyer2016summarizing}. 
However, these models were often task-specific and were trained on particular datasets, limiting their adaptability to custom tasks.
Recently, with the increasing popularity of pre-trained models in natural language processing, similar pre-trained models for code have been introduced, including  CodeBERT \cite{feng2020codebert}, GraphCodeBERT \cite{guo2020graphcodebert}, PLBART \cite{ahmad2021unified}, CodeT5 \cite{wang2021codet5}, UniXCoder \cite{guo2022unixcoder}, CodeGen \cite{nijkamp2022codegen}, and StarCoder \cite{li2023starcoder}.

\paragraph{Incorporating Structured Representations for Code}
Given that code can be represented by structures such as syntax or parse trees, data flows, and execution flows, several efforts explored various approaches to integrate these structures into code models.
For instance, \cite{mou2016convolutional} explored a convolutional network applied to the program's AST, while 
\cite{phan2017convolutional} proposed leveraging the program's execution flow in combination with graph-based CNNs for software defect prediction.
Meanwhile, \cite{hu2018deep} proposed converting an AST into a sequence using tree traversal and employing an LSTM to generate code comments.
\cite{zhou2019devign} utilized composite code representations including the AST, control flow, data flow, and natural code sequence for vulnerability detection.
To incorporate code structures into the Transformer architecture, prior efforts introduced novel positional encodings that represent node positions within the tree \cite{shiv2019novel,peng2022rethinking}, or utilized syntax or parse tree traversals or path information to provide structural information to the model \cite{kim2021code,wang2021code,peng2021integrating,jiang2021treebert,guo2020graphcodebert,Tipirneni2022}.
Specific tree-based attention mechanisms have also been proposed \cite{Chen2018,tang2022ast,10123596} and non-learnable parse-tree encoders were used in conjunction with sequence Transformers to interpret how syntax is used in downstream tasks \cite{Zanzotto2020}.

%% file: sections/method.tex
\section{Structured Code Representation}
A compiler transforms the high-level source code written by programmers into low-level machine code that can be executed on a computer's hardware.
Broadly speaking, the compilation consists of two phases: the front end (analysis) and the back end (synthesis) \cite{aho2007compilers}.
The front end includes lexical analysis (tokenization), syntax analysis (parsing), and semantic analysis, generating an AST or a similar intermediate representation that captures the program's logical structure and semantics. 
A grammar  specifies how keywords, identifiers, and literals should be structured, which the front end uses to break the source code into individual tokens and create the syntax tree, ensuring that the source code adheres to the language's syntax rules.

\subsection{Concrete Syntax Tree}\label{sec:CST}
The compiler front end typically starts by creating a CST from the input source code and then abstracts away the non-essential details from the CST to create the AST.
A CST is a tree representation of the source code according to the language's grammar.
It closely mirrors the code's textual structure and includes all the syntactic details like parentheses and punctuation.
In contrast, an AST simplifies the CST, retaining only the program's logical structure.
While ASTs are more commonly used due to the balance they strike between compactness and preserving essential program structures for further analysis and transformations, we use CSTs for program representations mainly because they faithfully retain all the textual information, including punctuations, whitespaces, and formatting.
This preservation is essential when one must provide all the details of the code to the model and precisely reconstruct the code from the model-generated tree.
CSTs are more generally applicable than ASTs, as they can be built directly from the language's grammar without resorting to language-specific optimizations or implementations required by the ASTs.
We do note however that CSTs are typically more verbose than ASTs due to their inclusion of all syntactic details.

In practice, for code models all that is needed is a parser that converts a program to its syntax tree and an unparser that reverts the conversion.
Much of the related work \cite{Zhang2016,Dong2016,Rabinovich2017,Yin2017,Chen2018,Yin2018,Sun2019,Sun2020} works on ASTs because the unparsing support exists (for example, the Python \verb|ast| module offers both \verb|ast.parse()| and \verb|ast.unparse()|).
For a wide variety of languages, however, writing the unparser for each language is tedious, and thus we opt to use CSTs in this work instead.
It is worthwhile to note that in many recent papers that incorporate ASTs in code models, the syntax trees are in fact computed by Tree-sitter and these trees are CSTs.

\subsection{Serialization and Deserialization}\label{sec:serial}
A CST is an n-ary tree, where each node can have a finite but arbitrary number of children.
We call the leaf nodes, \emph{terminal nodes}, and the rest, \emph{non-terminal nodes}.
The terminal nodes contain the code text while the non-terminal nodes correspond to grammar rules.
In Figure~\ref{fig:cst-obj}, terminal nodes have a filled background; it is clear that these nodes are sufficient to deduce the program from the tree.

In order to use this representation with pre-trained models, one must serialize the tree to make it consumable by a sequence Transformer, which in turn admits a guaranteed process for deserialization in subsequent uses. To this end, we define serialization based on a hybrid pre-post-order tree traversal, where a node is visited before its ordered children and is revisited after all its descendants have been visited.
Let \verb|Node| denote the current node and let \verb|subtree| denote the subtree rooted at \verb|Node| and excluding \verb|Node|.
Serialization then reads (recursively) \verb|(_.Node subtree Node._)|, where \verb|(_.Node| and \verb|Node._)| are separate tokens that represent the first and second visit of \verb|Node| in the traversal.
Explicitly marking the boundary of the subtree by \verb|(_.Node| and \verb|Node._)| is necessary to ensure unambiguous deserialization.
See Figure~\ref{fig:cst-obj} for the serialization of a simplified tree and the Appendix for more examples of serialized CSTs in full.

\section{Adapting Pre-Trained Models to Program Structures}\label{sec:losses}
Existing pre-trained models are generally trained on source code treated as plain text without explicit incorporation of the structural information, which allows for easier pre-processing and more efficient pre-training (due to the sequence length being shorter).
We propose to adapt these models to code structures by using serialized parse trees for continual pre-training and fine-tuning.
Empirically, this approach demonstrates {high efficiency}: continual pre-training requires only a tiny fraction of time (i.e., number of steps) compared with the preexisting large-scale pre-training; and fine-tuning can achieve good performance on only a small number of labeled data.

While fine-tuning on supervised data with CSTs is relatively straightforward, how to continue pre-training on self-supervised objectives that make use of program structures is an open question.  We develop pre-training tasks that effectively learn the structural elements that can go in cohort with the remaining code and text.
In what follows, we use $\vx$, $\vy$, $\vz$ to denote natural language text, program text, and the corresponding serialized parse tree, respectively.

\subsection{Encoder-Decoder Models}

\paragraph{Masked SubTree Prediction (MSP)}
To learn the tree structure, we propose to randomly mask subtrees before encoding and ask the decoder to predict them (see Figure~\ref{fig:cst-obj}).
Specifically, we mask 15\% of the nodes in the parse tree, by randomly selecting non-terminal nodes and masking their entire subtrees, until the budget is attained.
We replace each of the selected subtrees with the same mask token and ask the model to predict these subtrees.
After serialization, masking subtrees is effectively equivalent to masking the corresponding span from the sequence of the serialized tree.
Hence, this technique can be considered a tree extension of the masked span prediction objective popularly used in natural language models \cite{raffel2020exploring,joshi2020spanbert,wang2021codet5}.
The training objective can be written as
\begin{equation}\label{eqn:MSP}
\mathcal L_{\text{MSP}} = \sum_{t=1}^k \log p(\vz_t^{\text{mask}} | \vz^{\backslash \text{mask}}, \vz_{<t}^{\text{mask}}),
\end{equation}
where $\vz^{\backslash \text{mask}}$ is the masked input and $\vz_t^{\text{mask}}$ is the masked sequence to predict with length $k$.
Since the subtrees include non-terminal and terminal nodes, this task aids the model in learning the relationship between language grammar (non-terminal nodes) and code semantics (terminal nodes).

\paragraph{Masked Node Prediction (MNP)}
To enable the model to further learn the language grammar, we propose the masked node prediction objective, where all the non-terminal nodes are masked by a unique sentinel token and they together form, in the original order, a sequence $\mI$.
The decoder is then asked to predict $\mI$ (see Figure~\ref{fig:cst-obj}).
This task is similar to the masked identifier prediction objective of \cite{wang2021codet5} and the DOBF objective of \cite{lachaux2021dobf}, except that the masked tokens here are not the original program tokens.
The training objective can be written as
\begin{equation}\label{eqn:MNP}
\mathcal L_{\text{MNP}} = \sum_{t=1}^{|\mI|} \log p(\mI_t | \vz^{\backslash \mI}, \mI_{<t}),
\end{equation}
where $\vz^{\backslash \mI}$ denotes the serialized tree with non-terminal nodes masked.
This is a much harder task than the MSP objective because in MNP the majority of the tree is masked and the model has to learn the language grammar to predict them.

\paragraph{Text-to-Tree (TeTr) and Tree-to-Text Conversion (TrTe)}
When the data contains natural language $\leftrightarrow$ code pairs (e.g., code with the natural language description), we encode either part and decode the other to align them.
The training objectives can be written as
\begin{equation}\label{eqn:TeTr}
\begin{aligned}
    \mathcal L_{\text{TeTr}} &= \sum_{t=1}^{|\vz|} \log p(\vz_t | \vx, \vz_{<t}),\\
    \mathcal L_{\text{TrTe}} &= \sum_{t=1}^{|\vx|} \log p(\vx_t | \vz, \vx_{<t}),
\end{aligned}
\end{equation}
which are similar to the typical objectives $\log p(\vy_t | \vx, \vy_{<t})$ and $\log p(\vx_t | \vy, \vx_{<t})$ used in \cite{wang2021codet5}, except that code $\vy$ is replaced by seralized tree $\vz$.
These two tasks match the downstream utilization of the model, where it has to either generate code given a natural language description or summarize the given code in natural language.

\subsection{Decoder-Only Models}
For decoder-only models, we find that it is effective to simply reuse the causal language modeling objective over the serialized tree $\vz$:
\begin{equation}
\mathcal L_{\text{DEC}} = \sum_{t=1}^{|\vz|} \log p(\vz_t | \vz_{<t}).
\end{equation}
When there exist natural language $\leftrightarrow$ code pairs, we replace $\vz$ in the above formula by $\vz' = [ \vx : \vz ]$; that is, concatenating text and code.
Compared with the specialized objectives of MSP and MNP for encoder-decoder models, here we require the model to reconstruct all tokens (both terminal and non-terminal nodes) in an autoregressive manner.

\section{Experimental Setup}

\paragraph{Pre-Trained Models}
We use \textbf{CodeT5} \cite{wang2021codet5} as our encoder-decoder pre-trained model, and  \textbf{CodeGen} \cite{nijkamp2022codegen} as our pre-traaned decoder-only model.
The majority of the experiments were conducted on CodeT5-base (220M) and CodeGen-Multi (350M), but we also include results with progressively larger CodeT5 models to study the effect of scale.

\paragraph{Tokenizer}
To adapt these pre-trained models to code structures, we augment the respective tokenizer by including the non-terminal nodes, each treated as a single token.
This allows the model to learn targeted embeddings for the non-terminal nodes and also makes the input/output length manageable.
Later, we show that not including the non-terminal node tokens in the tokenizer substantially degrades downstream performance.

\paragraph{Pre-Training Dataset}
We use CodeSearchNet \cite{husain2019codesearchnet} augmented partially by the Stack \cite{kocetkov2022stack} for continual pre-training on structured self-supervised objectives.
CodeSearchNet contains nearly 6.5 million data samples extracted from the most popular GitHub projects with permissive licenses.
Among them, around 2.3 million samples have comments accompanying the code.
CodeSearchNet contains six programming languages; however, it misses some languages needed in our downstream tasks (namely, C and C\#).
To include them, we subsample 1 million samples for each from Stack, which results in a total of 8.5 million samples across eight programming languages.
More details are in the Appendix. %

\paragraph{Continual Pre-Training}
We further pre-train CodeT5 and CodeGen by using the objectives described in Section~\ref{sec:losses}.
We train for one epoch, by using 32 V100-32GB GPUs with a batch size of 1024.
Hyper-parameters are in the Appendix. %
Note that continual pre-training is rather lightweight.
Compared with the reported training time of 12 days by using 16 A100-40GB GPUs for CodeT5 and 450,000 steps for CodeGen, our continual pre-training takes only 15 hours for CodeT5 and approximately 8,000 steps for CodeGen.

\paragraph{Data-Efficient Fine-Tuning}
Existing benchmarks for evaluating code models typically use datasets containing a few thousand to several hundred thousand data samples.
For instance, the Java $\leftrightarrow$ C\# translation dataset in CodeXGLUE \cite{DBLP:journals/corr/abs-2102-04664} contains 10.3k samples.
While it is possible to curate data at expensive cost in a one-time effort, often dataset creation can be financially burdensome and even practically unattainable.
This issue is even more pronounced for low-resource languages or domain-specific languages.
Hence, we focus on low-data scenarios in the experiments.

\paragraph{Evaluation Metrics}
Following the literature, we use four metrics: the BLEU score, typically used to evaluate text generation \cite{papineni2002bleu};
the CodeBLEU score, which addresses BLEU's weak correlation with semantic code correctness \cite{ren2020codebleu};
Exact Match (EM);
and pass@$k$, which addresses EM's weakness in capturing code variability and is an unbiased estimator of functional correctness \cite{chen2021evaluating}.

%% file: sections/experiment.tex
\section{Results}

\begin{figure*}[ht]
  \includegraphics[width=0.24\textwidth]{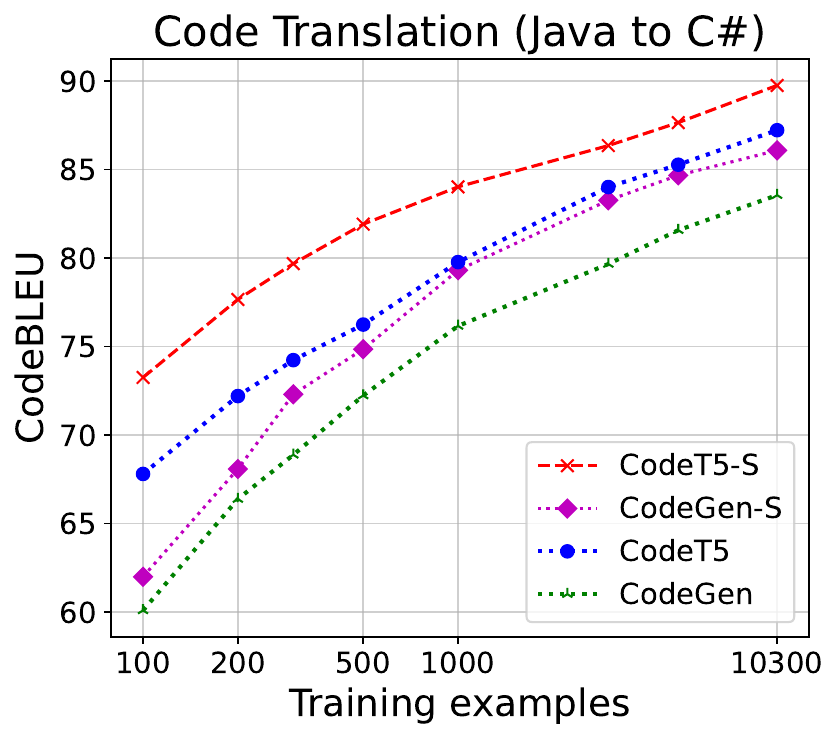} \hfill
  \includegraphics[width=0.24\textwidth]{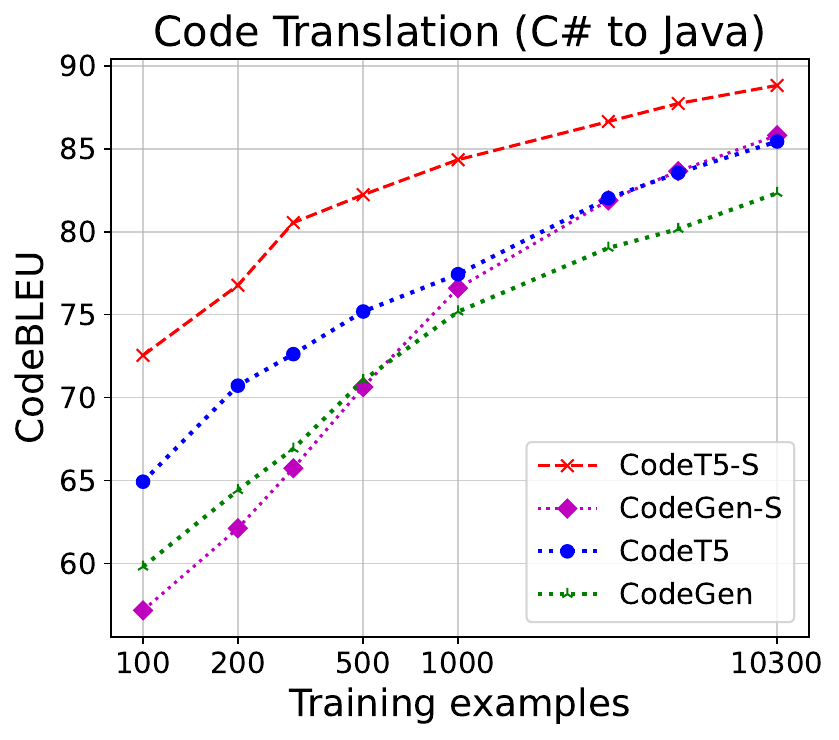} \hfill
  \includegraphics[width=0.24\textwidth]{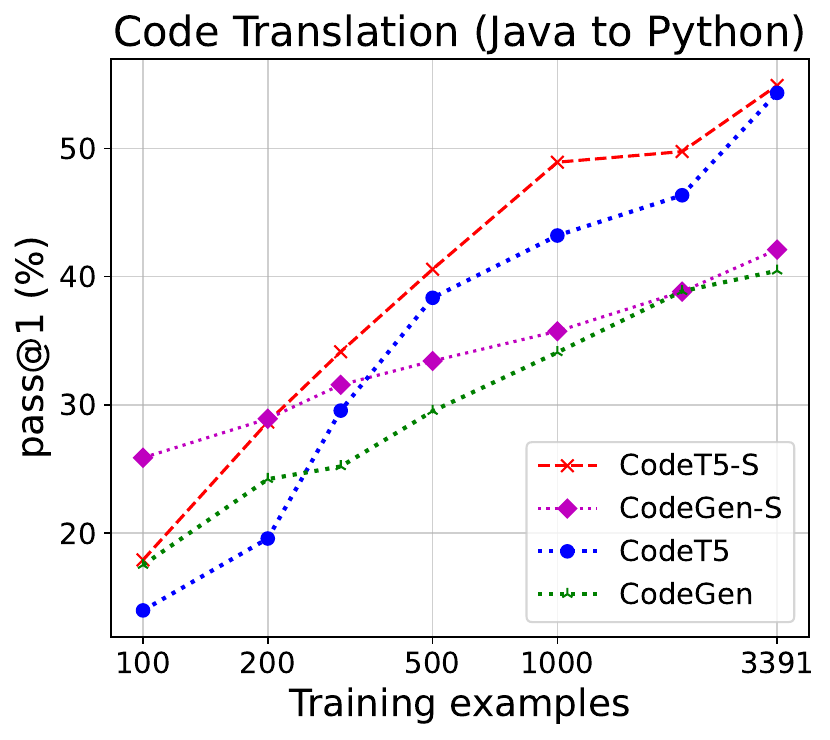} \hfill
  \includegraphics[width=0.24\textwidth]{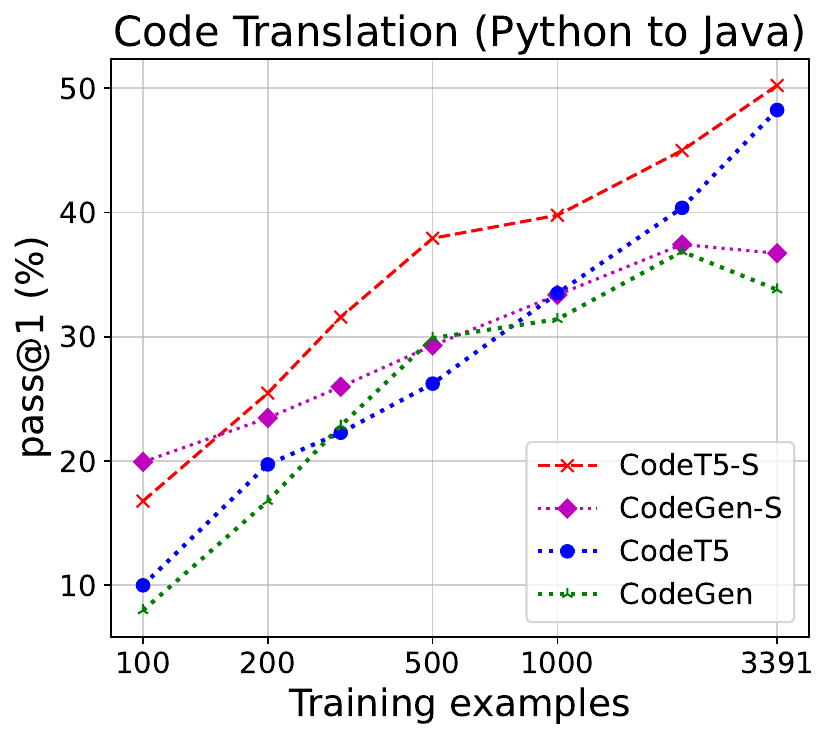}
  \caption{Code translation performance. Left two: Java $\leftrightarrow$ C\# (CodeXGLUE); right two: Java $\leftrightarrow$ Python (TransCoder). For full results on all evaluation metrics, see the Appendix.}
  \label{fig:codetrans}
\end{figure*}

\begin{figure*}[ht]
  \begin{minipage}[t]{0.74\textwidth}
  \includegraphics[width=0.32\textwidth]{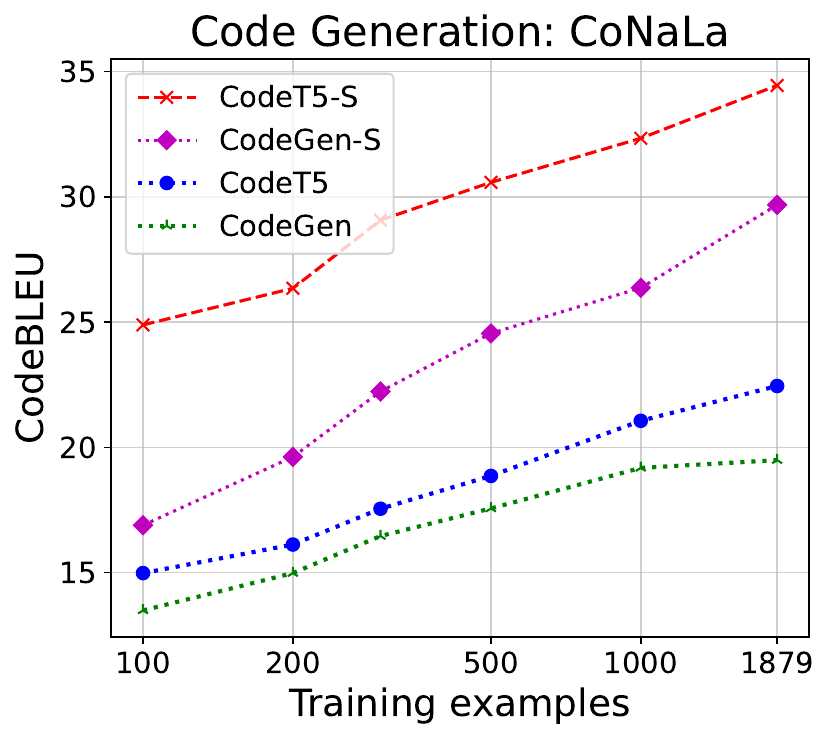} \hfill
  \includegraphics[width=0.32\textwidth]{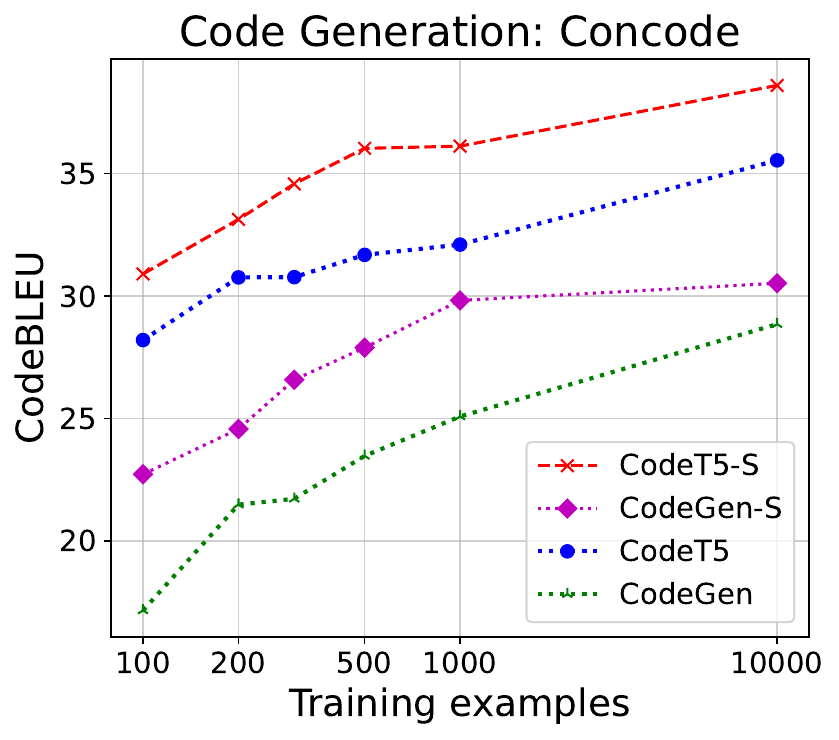} \hfill
  \includegraphics[width=0.32\textwidth]{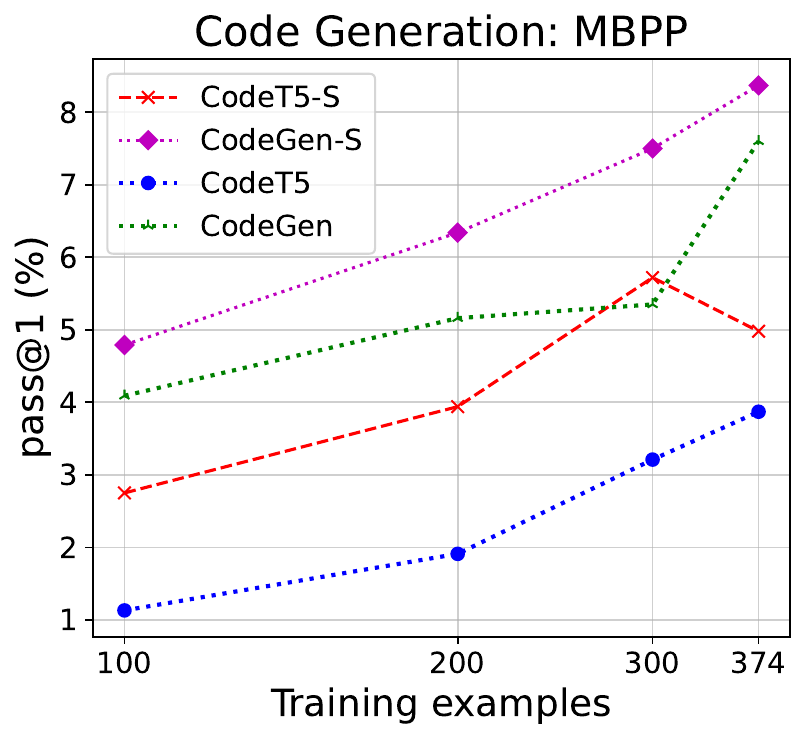}
  \caption{Code generation performance. From left to right: CoNaLa, Concode, MBPP. For full results on all evaluation metrics, see the Appendix.}
  \label{fig:text2code}
  \end{minipage} \hfill %
  \begin{minipage}[t]{0.24\textwidth}
  \includegraphics[width=\textwidth]{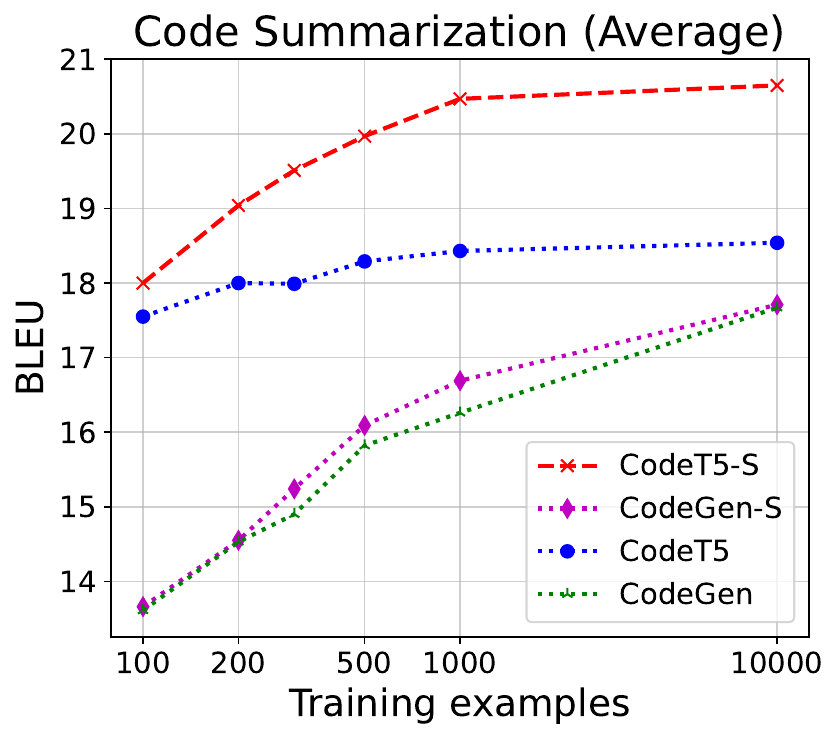}
  \caption{Average Code Summarization performance. For each language, see the Appendix.}
  \label{fig:code2text}
  \end{minipage}
\end{figure*}

\paragraph{Code Translation (Code2Code)}

We experiment with two datasets: CodeXGLUE (Java $\leftrightarrow$ C\#) \cite{DBLP:journals/corr/abs-2102-04664} and 
TransCoder (Java $\leftrightarrow$ Python) \cite{roziere2020unsupervised}.
Since the TransCoder dataset only provides the test samples, we use the AVATAR dataset \cite{Ahmad2023} as the training (fine-tuning) set for Java $\leftrightarrow$ Python translation.
Figure~\ref{fig:codetrans} shows typical results (averaged over 3 random seeds) for both datasets and all translation directions, with more results on other evaluation metrics left to the Appendix due to page limitation.
We observe that the structured model (suffixed with -S) almost always outperforms the base model.
Encouragingly, for CodeXGLUE, both structured models achieve a comparable performance by using only 3k training examples, to the counterpart base models by using the full training set (10.3k examples).
Moreover, with only 100 training examples, CodeT5-S outperforms the base counterpart by at least 5 CodeBLEU points on the CodeXGLUE dataset and CodeGen-S outperforms the base counterpart by at least 10 percent more passing programs on the TransCoder dataset.

\paragraph{Code Generation (Text2Code)}

We experiment with three datasets: CoNaLa \cite{yin2018mining}, Concode \cite{iyer2018mapping}, and MBPP \cite{austin2021program}.
We highlight that for all three tasks -- and especially for the MBPP dataset -- we model the problem as a strict Text2Code task, where the input is the natural language description and the output is the complete code.
This differs from MBPP task formulations in prior works, where the input is the natural language description and the Python function signature and docstring and the model generates the function body.
Figure~\ref{fig:text2code} shows typical results (averaged over 3 random seeds) for all of them, with more results on other evaluation metrics left to the Appendix.
We observe that the structured models always outperform the base counterparts.
On CoNaLa, CodeT5-S achieves a higher CodeBLEU score by using only 100 training examples, compared with CodeT5 using over 1.8k examples.
On Concode, the performances of both structured models by using 500 examples are comparable to those of the base counterparts using 10k examples (note that the full training set contains over 100k examples, but we focus on the low-data scenarios for comparison only).

\paragraph{Code Summarization (Code2Text)}

We experiment with the CodeSearchNet \cite{husain2019codesearchnet} dataset, which contains six languages.
Figure~\ref{fig:code2text} shows the average performance, with the results on each language left to the Appendix.
We observe that the structured model CodeT5-S significantly improves over the base model CodeT5.
The performance of CodeT5 saturates over 1k training examples, while CodeT5-S can achieve this performance with fewer than 200 examples.
Moreover, when both models use 1k training examples, CodeT5-S outperforms CodeT5 by 2 BLEU points.
Even though we do not observe a similarly significant improvement over CodeGen, the structured model CodeGen-S is competitive and performs slightly better.

\paragraph{Ablation Analysis}

For all the above tasks, we investigate how the structured model obtains superior performance by upgrading the base counterpart step by step.
Table~\ref{tab:ablation-scores} suggests that simply fine-tuning the base model on serialized parse trees results in a performance drop.
The main shortcoming is that the tokenizer may tokenize the decorated text of the non-terminal nodes (i.e., a word prepended by \verb|(_.| or appended by \verb|._)|) into multiple parts and forfeit the structural meaning of these nodes.
Then, adding non-terminal tokens to the tokenizer gains back the performance.
Still, for Code2Text, the performance is not as competitive as the base model.
Finally, using continual pre-training (which leads to our structured model) yields additional gains and beats the base model by a large margin.

Additionally, we investigate the importance of the proposed training objectives for encoder-decoder models.
Table~\ref{tab:ablation-scores} shows four scenarios, each using only one or two of the objectives for training.
The performance of CodeT5-S suffers in all these scenarios, but the best among them is still better than the base CodeT5, corroborating the advantage of leveraging structures.

\begin{table}[t]
  \caption{Ablation analysis (low-data scenario).}
  \label{tab:ablation-scores}
  \centering
  \resizebox{.99\linewidth}{!}{
  \begin{tabular}{clccc}
    & & \textbf{Code2Code} & \textbf{Text2Code} & \textbf{Code2Text} \\
    & Dataset $\rightarrow$ & Java $\rightarrow$ C\# & CoNaLa & Go \\
    & Metric $\rightarrow$ & CodeBLEU & CodeBLEU & BLEU \\
    & Train size $\rightarrow$ & 500 & 500 & 500\\
    \midrule
    \multirow{4}{*}{\rotatebox[origin=c]{90}{\parbox[c]{1.5cm}{\centering Enhancing CodeT5}}}
    & CodeT5            & 76.24 & 18.86 & 17.51 \\
    & + Trees           & 65.04 & 3.94  & 16.03 \\
    & + Tokenizer       & 77.83 & 25.99 & 16.69 \\
    & \textbf{CodeT5-S} & \textbf{81.91} & \textbf{30.58} & \textbf{24.49} \\
    \midrule
    \multirow{4}{*}{\rotatebox[origin=c]{90}{\parbox[c]{1.5cm}{\centering Training objectives}}}
    & $\triangleright$ Only MSP    & 81.35 & \underline{29.53} & 17.49 \\
    & $\triangleright$ Only MNP    & 74.51 & 24.18 & 16.15 \\
    & $\triangleright$ TeTr + TrTe & 76.40 & 29.27 & \underline{22.28} \\
    & $\triangleright$ MSP + MNP   & \underline{81.49} & 28.49 & 17.45 \\
  \end{tabular}
  }
\end{table}

\paragraph{Comparison With Other Code Models}

We compare the performance of our structured model, CodeT5-S, on all tasks, with that of three representative code models:
PLBART \cite{ahmad2021unified}, which is not structure-aware; UniXCoder \cite{guo2022unixcoder}, which leverages syntax trees but focuses on pre-training rather than adaptation; and StructCoder \cite{Tipirneni2022}, which exploits not only syntax trees but also data flow graphs.
All three models are of comparable sizes and pre-trained with a similar amount of data - except PLBART which is trained with roughly one order of magnitude more tokens than the other models.
Table~\ref{tab:compare.other.models} shows that our model achieves the best performance across the board.
Not surprisingly, the model not exploiting structures (PLBART) performs the worst.
For structured models, UniXCoder uses the non-terminal nodes of the syntax tree in pre-training but not fine-tuning; while StructCoder does not serialize the syntax tree as we do but rather, uses the root-leaf paths to provide embeddings for the terminal nodes.
In contrast, our approach is much simpler, makes use of the non-terminal nodes without changing the architecture of the pre-trained model, and performs better empirically.

\begin{table}[t]
  \caption{Performance comparison with other code models.}
  \label{tab:compare.other.models}
  \centering
  \resizebox{.99\linewidth}{!}{
  \begin{tabular}{lccc}
    & \textbf{Code2Code} & \textbf{Text2Code} & \textbf{Code2Text} \\
    Dataset $\rightarrow$ & Java $\rightarrow$ C\# & CoNaLa & Go \\
    Metric $\rightarrow$ & CodeBLEU & CodeBLEU & BLEU \\
    Train size $\rightarrow$ & 100 & 100 & 100 \\
    \midrule
    \textbf{CodeT5-S (220M)} & \textbf{72.79} & \textbf{23.86} & \textbf{22.93} \\
    PLBART (140M) & 66.28 & 12.43 & 12.57 \\
    UniXCoder (125M) & 71.38 & 12.92 & \underline{18.02} \\
    StructCoder (220M) & \underline{72.10} & \underline{20.28} & 11.22 \\
  \end{tabular}
  }
\end{table}

\paragraph{Different Model Sizes}

We investigate the performance of structured adaptation under the influence of model sizes.
Table~\ref{tab:scaling} shows the results of CodeT5-S and the improvement over CodeT5 (annotated inside parentheses).
We observe that the structured model sustains noticeable improvement over the base model across the board.
More importantly, the improvement does not diminish for larger models, indicating that structures always complement text-based training for code.
We note that in some cases, the performance is not always better on a larger model (see the 770M case), but similar observations are made in other publications as well (see, e.g., Table 9 of \cite{Le2022}), which likely is caused by some artifacts of fine-tuning, especially with limited training data.

\begin{table}[t]
  \caption{Quality and improvement (in parentheses) of CodeT5-S over CodeT5 on different model sizes.}
  \label{tab:scaling}
  \centering
  \resizebox{.99\linewidth}{!}{
  \begin{tabular}{lccc}
    & \textbf{Code2Code} & \textbf{Text2Code} & \textbf{Code2Text} \\
    Dataset $\rightarrow$ & Java $\rightarrow$ C\# & CoNaLa & Go \\
    Metric $\rightarrow$ & CodeBLEU & CodeBLEU & BLEU \\
    Train size $\rightarrow$ & 100 & 100 & 100 \\
    \midrule
    60M  & 71.01 (+6.19) & 20.87 (+6.96) & 19.09 (+2.89) \\
    220M & 72.79 (+6.07) & 23.86 (+9.33) & 22.93 (+5.37) \\
    770M & 71.91 (+4.97) & 22.49 (+7.13) & 25.10 (+9.50)   \\
  \end{tabular}
  }
\end{table}

\paragraph{Qualitative Analysis}

We provide several examples to qualitatively demonstrate the superior code generation quality (on the MBPP dataset) and code translation quality (on the AVATAR dataset) of our structured models.
Figure~\ref{fig:mbpp_example_3} gives one example, which shows that only the structured models generate recursive functions (as list elements may be lists recursively).
More examples are given in the Appendix.
These examples suggest that the base models may produce programs that either do not follow instructions, alter the ordering of list items, are syntactically incorrect, or do not match the source programs closely and thus produce incorrect test results.
In contrast, the programs produced by our structured models avoid these mistakes.

\input{code_examples/mbpp_example_3}

\paragraph{Error Breakdown}

\begin{figure}[ht]
  \centering
  \includegraphics[width=\linewidth]{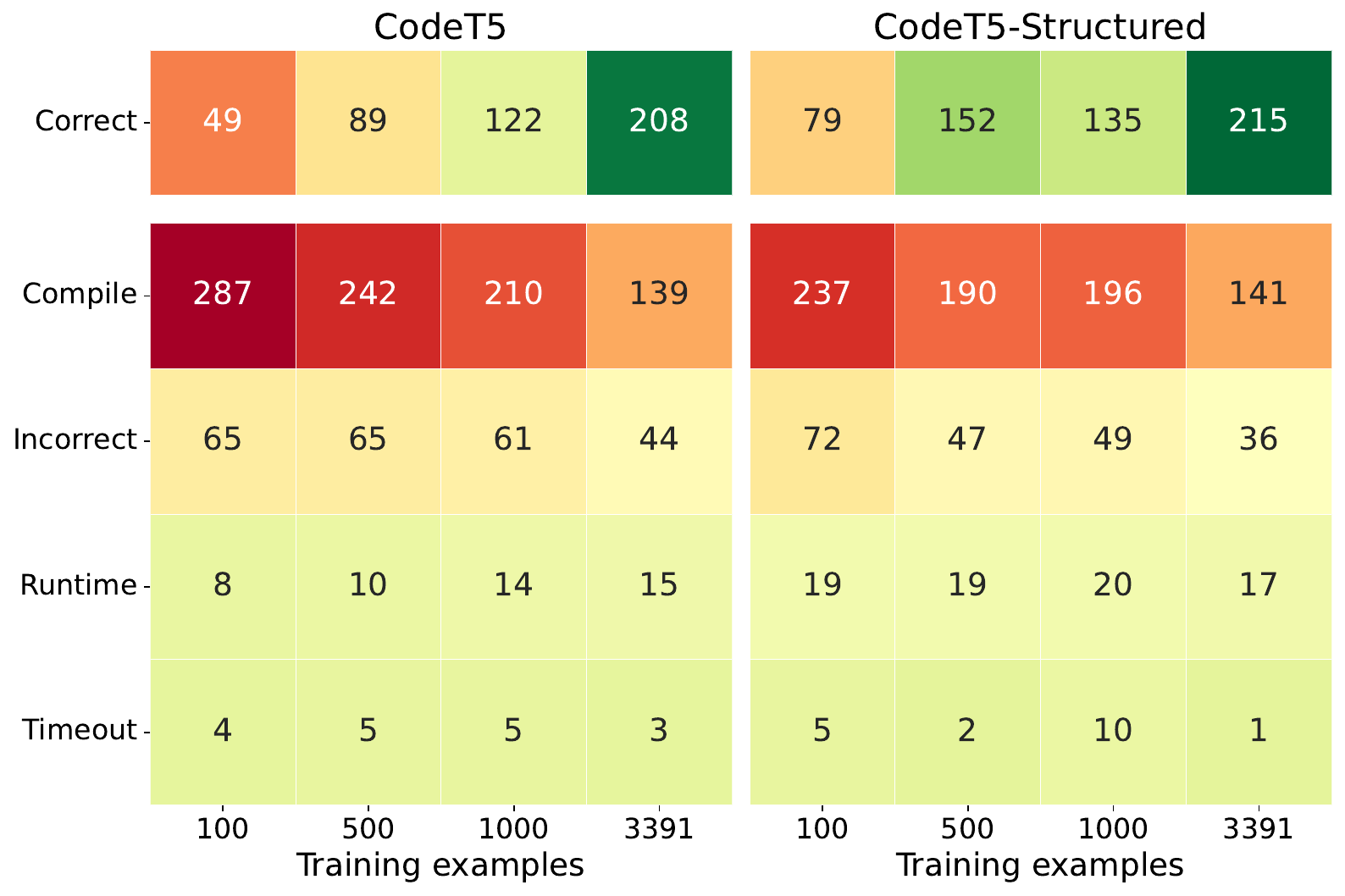}
  \caption{Analysis of Python $\rightarrow$ Java code translation results. Incorrect translations are categorized by errors: compilation error (Compile), incorrect answer (Incorrect), runtime error (Runtime), and execution timeout (Timeout).}
  \label{fig:error.analysis}
\end{figure}

We investigate the performance gains of the structured model over the base model, through a breakdown of correct and incorrect programs, using the TransCoder (Python $\to$ Java) dataset.
Figure~\ref{fig:error.analysis} shows that with more training examples, the number of correct translations generally increases while the various errors decrease.
Notably, the largest decrease occurs in compilation errors (compared with incorrect results, runtime errors, and timeouts).
This finding corroborates the advantage of introducing parse structures into the code model, which enhances its translation quality by conforming to syntax rules.

\paragraph{Code Complexity}

\begin{figure}[ht]
  \centering
  \includegraphics[width=0.8\linewidth]{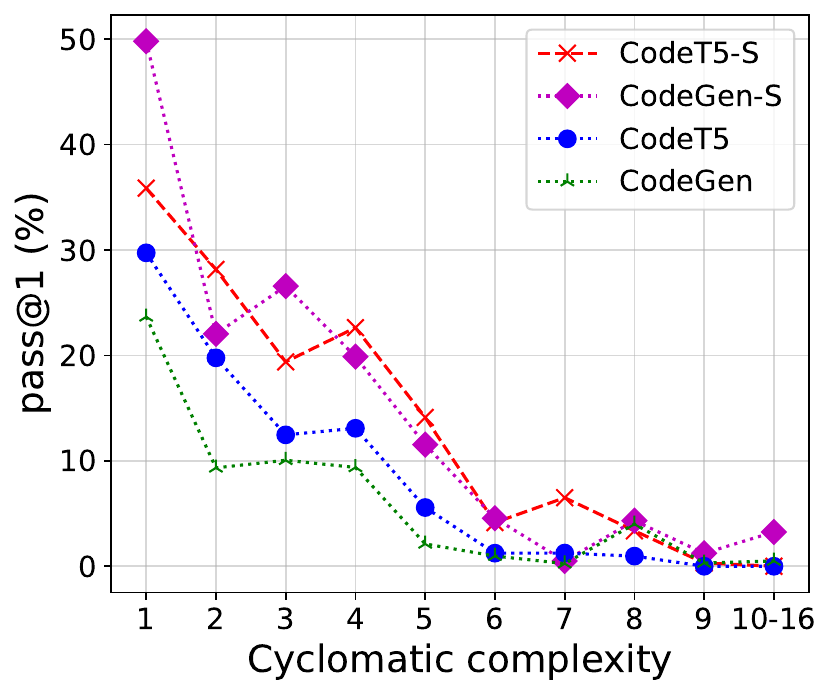}
  \caption{Code translation performance versus input code complexity on the Python $\rightarrow$ Java translation task.}
  \label{fig:complexity}
\end{figure}

We further investigate how the translation quality varies as the input programs become increasingly more complex, as measured by Cyclomatic complexity \cite{mccabe1976complexity}. %
Figure~\ref{fig:complexity} plots results using the TransCoder (Python $\to$ Java) dataset with models trained using 100 examples.
It suggests that as the input Python program becomes more complex, the percentage of correct Java translations generally decreases as expected.
Importantly, for each complexity level, the structured models always perform better than the base counterparts.
We provide some translation examples of complex programs in the Appendix.
In these examples, the base models make mistakes in the syntax or semantics, but our structured models faithfully translate the source program.

%% file: code_examples/mbpp_example_3.tex
\begin{figure*}[ht]
\begin{tcolorbox}[toprule at break=0pt, bottomrule at break=0pt, left=0.4em, right=0.4em, top=0em, bottom=0.2em]
\begin{lstlisting}[basicstyle=\fontsize{9}{10}\ttfamily]
(*@\codehl{Write a function to flatten a given nested list structure.}@*)
\end{lstlisting}
\begin{tcbitemize}[raster columns=2, raster equal height=rows, raster force size=false, left=0.4em, right=0.4em, top=0em, bottom=0em]
\tcbitem[title=CodeT5]
\begin{lstlisting}[language=Python]
def flatten_list(l):
  result = [l]
  for m, n, o in zip(l):
    if (m == n == o):
      result.append(m)
  return result
\end{lstlisting}
\tcbitem[title=CodeGen]
\begin{lstlisting}[language=Python]
def flatten_list(nested_list):
  return [item for sublist in nested_list \
          for item in sublist]
\end{lstlisting}
\tcbitem[title=CodeT5-Structured]
\begin{lstlisting}[language=Python]
def flatten_list(l):
  result = [ ]
  for ele in l:
    if isinstance(ele, list):
      result.extend(flatten_list(ele))
    else:
      result.append(ele)
  return result
\end{lstlisting}
\tcbitem[title=CodeGen-Structured]
\begin{lstlisting}[language=Python]
def flatten_list(l):
  res = [ ]
  for i in l:
    if isinstance(i, list):
      res.extend(flatten_list(i))
    else:
      res.append(i)
  return (res)
\end{lstlisting}
\end{tcbitemize}
\end{tcolorbox}
\caption{Model generations for a test sample in the MBPP dataset.}
\label{fig:mbpp_example_3}
\end{figure*}

%% file: sections/conclusion.tex
\section{Conclusion}
In this work, we explore data-efficient adaptation of code language models by utilizing the serialized parse tree of source code.
We develop training objectives for both encoder-decoder and decoder-only models to enable the continual pre-training of existing models and the adaptation to structures.
Evaluating the approach on multiple downstream tasks, including code translation, generation, and summarization, we observe significant gains in both fuzzy metrics and functional metrics, especially in the low-data regime where fine-tuning examples are limited.

%% file: sections/appendix.tex
\section{Serialized CST Examples}
\label{apdx:cst-serialization}

The main text gives an example of a simplified CST for visualization purposes.
Here, we show examples of two full CSTs.
See Figure~\ref{fig:code-ex}.
The first example is a Python program and the second example is a Go program.
We show the serialized trees in a pretty-printed format rather than a sequence format to facilitate reading.
Note that some languages (such as Python) use newlines and indentation to separate statements and structure code blocks.
For these languages, the CST explicitly includes the newline/indent/dedent nodes, which can unambiguously reconstruct the program text.

\section{Pre-Training Dataset Statistics}
\label{apdx:pretraining-datastats}

In Table \ref{datastats-pretrain}, we provide the pre-training dataset statistics.
We utilize 8.5 million data samples for continual pre-training, of which, 2.3 million have natural language annotation along with the code.
The remaining 6.2 million data samples have only the code.

\begin{table}[ht]
  \caption{Pre-Training data statistics.}
  \label{datastats-pretrain}
  \centering
  \begin{tabular}{cccc}
    \multicolumn{1}{c}{\bf Language} & \bf With NL & \bf W/O NL & \multicolumn{1}{c}{\bf Total} 
    \\ \midrule
    Go & 347,665 & 379,103 & 726,768 \\
    Ruby & 53,497 & 110,551 & 164,048 \\
    Python & 499,055 & 657,030 & 1,156,085 \\
    Java & 499,618 & 1,070,271 & 1,569,889 \\
    JavaScipt & 139,902 & 1,717,933 & 1,857,835 \\
    PHP & 579,763 & 398,058 & 977,821 \\
    C & 0 & 1M & 1M \\
    C\# & 0 & 1M & 1M
    \\ \midrule
    Total & 2,119,500 & 6,332,946 & 8,452,446
  \end{tabular}
  \vskip -10pt
\end{table}

\section{Hyper-parameters}
\label{apdx:hyperparams}

In Table \ref{tab:hyperparams-pretrain}, we provide the list of hyper-parameters for continual pre-training.
Additionally, we pre-train the models in \verb|fp16| precision.
For CodeT5, we set the input and output sequence lengths to a maximum of 512 tokens.
For CodeGen, we use a context length of 1024 tokens.
In Table \ref{tab:hyperparams-finetune}, we provide the list of hyper-parameters used for fine-tuning.
For MBPP and TransCoder results, we sample generations from the model using temperature = 0.8.

\begin{table}[ht]
  \caption{Hyper-parameters used for continual pre-training.}
  \label{tab:hyperparams-pretrain}
  \centering
  \begin{tabular}{ll}
    \multicolumn{1}{c}{\bf Hyperparameter}  &\multicolumn{1}{c}{\bf Value}
    \\ \midrule
    Learning Rate & 2e-4 \\
    No. of epochs & 1 \\
    Batch Size & 1024 \\
    Warmup steps & 750 \\
    Learning Rate Schedule & cosine
  \end{tabular}
\end{table}

\begin{table}[ht]
  \caption{Hyper-parameters used for fine-training.}
  \label{tab:hyperparams-finetune}
  \centering
  \begin{tabular}{ll}
    \multicolumn{1}{c}{\bf Hyperparameter}  &\multicolumn{1}{c}{\bf Value}
    \\ \midrule
    Learning Rate & 5e-5 \\
    No. of epochs & 100 \\
    Batch Size & 8 \\
    Warmup steps & 100 \\
    Learning Rate Schedule & cosine \\
    Early Stopping Patience & 5
  \end{tabular}
\end{table}

\section{Qualitative Examples}
\label{apdx:qualitative-examples}

In addition to the example shown in the main text, in Figures~\ref{fig:mbpp_example_1}--\ref{fig:avatar_example_2} we provide extra code generation examples for the MBPP test set and code translation examples for the TransCoder test set.
All models were fine-tuned with only 100 training examples.

In Figure~\ref{fig:mbpp_example_1}, the base models fail to follow the instruction that requires using lambda expressions, whereas our structured models succeed.

In Figure~\ref{fig:mbpp_example_2}, the use of \verb|sort()| in CodeT5's generation and \verb|set()| in CodeGen's generation inadvertently changes the ordering of items in the input list.
In comparison, the generations from our structured models avoid this mistake.

In Figure~\ref{fig:avatar_example_1}, CodeT5 incorrectly defines indexes in loops, misses out a condition in the \verb|if| statement, and produces a spurious closing curly bracket before the return statement, causing compilation error; while CodeGen produces a functionally incorrect program which looks very different from the source program.
In comparison, the translations from our structured models are correct and match the source Python code nearly line-by-line.

In Figure~\ref{fig:avatar_example_2}, CodeT5 uses the variables \verb|a| and \verb|b| without defining them, incorrectly initializes the value of the \verb|result| variable, and mistakenly places \verb|a += 1| inside the wrong \verb|if| block. CodeGen, on the other hand, defines extra variables not defined in the original program and computes a value from either the array \verb|A| or \verb|B|, rather than computing the smallest difference between \verb|A| and \verb|B|.
The translations from our structured models are again correct and match the source Python code nearly line-by-line.

\section{Code Translation Results}
\label{apdx:code-translation-alldatasets}

\begin{figure*}[ht]
    \centering
    \includegraphics[width=0.9\textwidth]{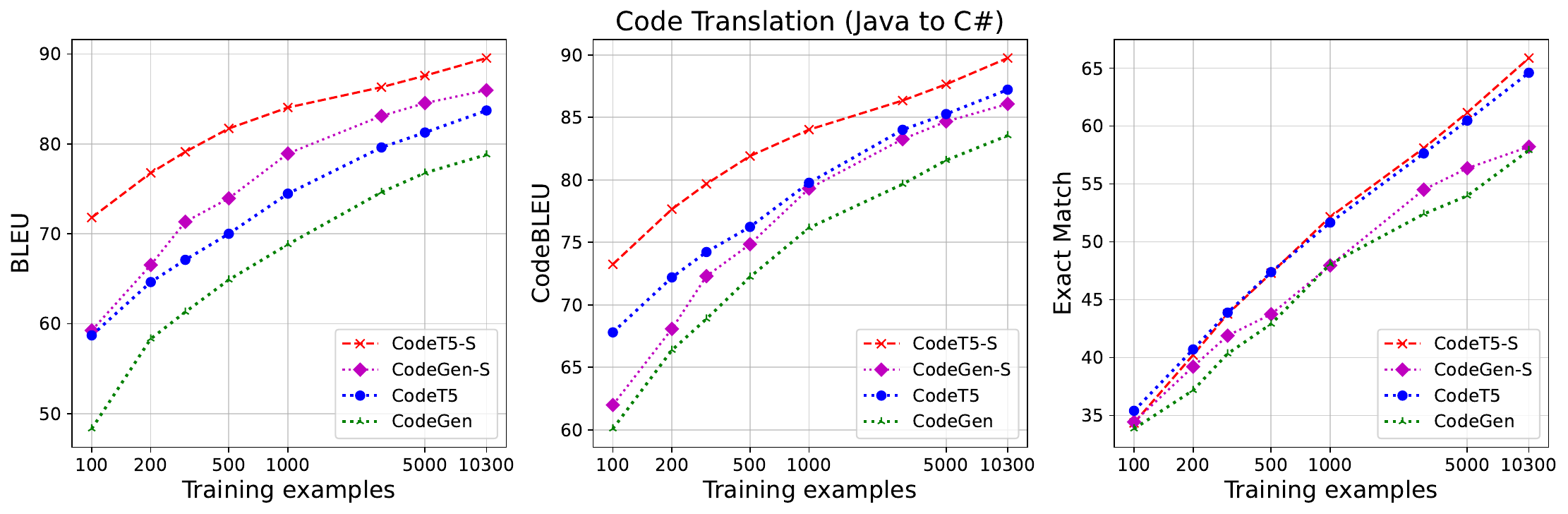}
    \\ \vspace{5pt}
    \includegraphics[width=0.9\textwidth]{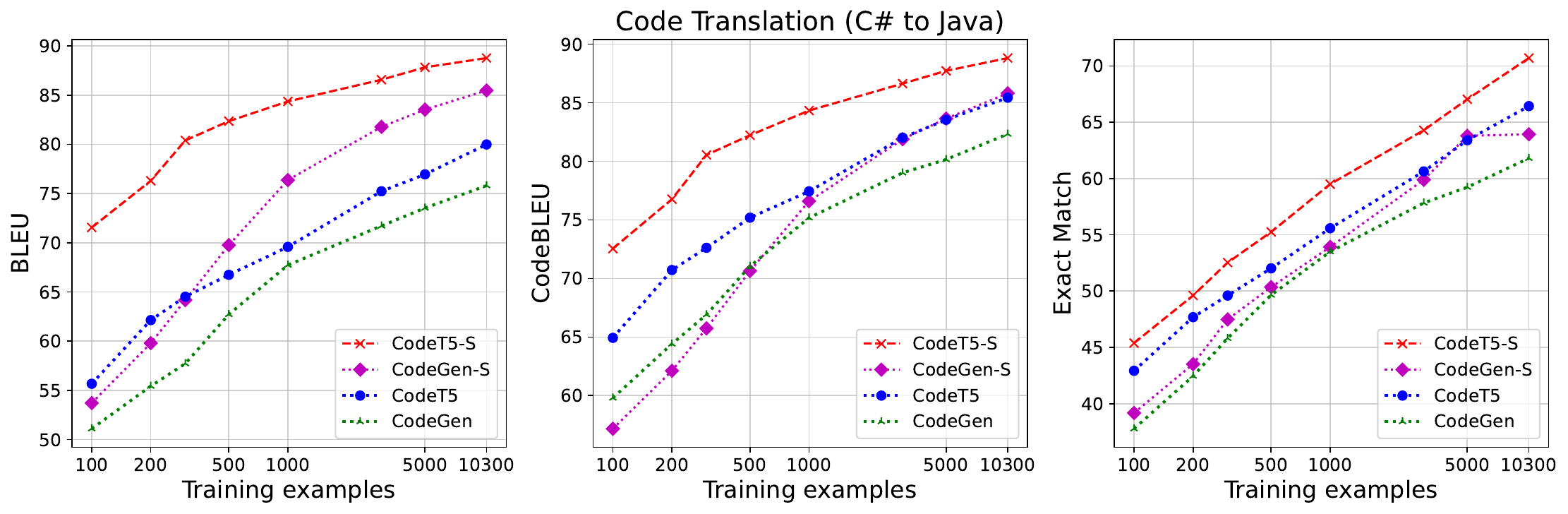}
    \\ \vspace{5pt}
    \includegraphics[width=0.9\textwidth]{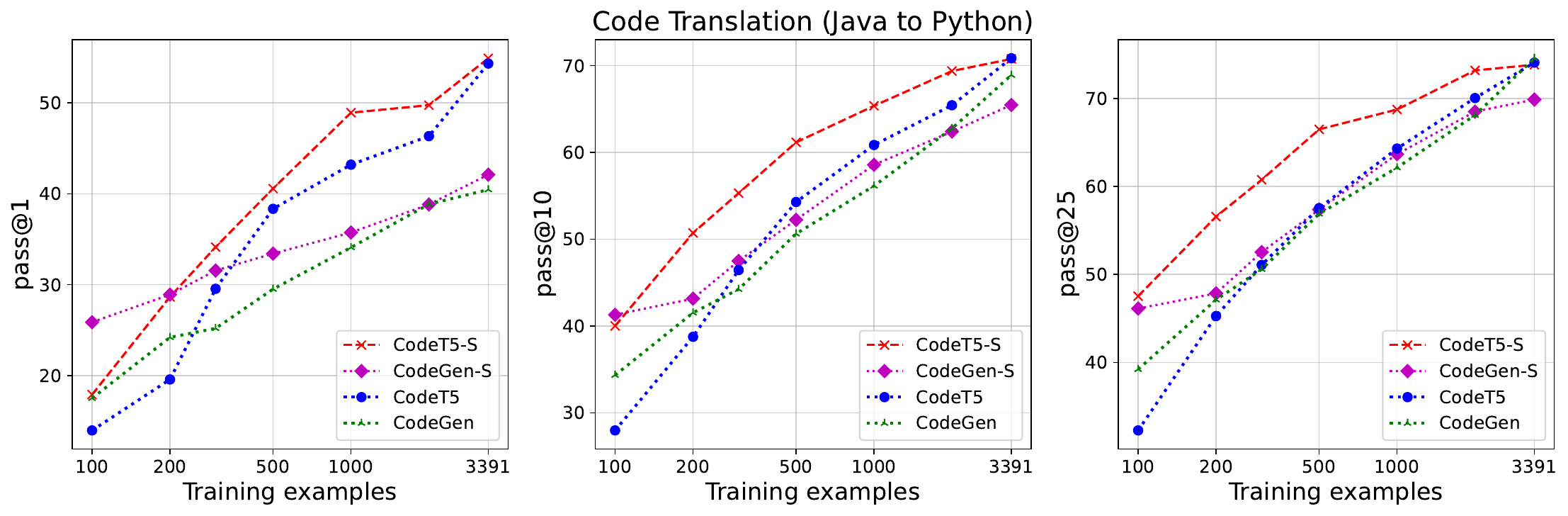}
    \\ \vspace{5pt}
    \includegraphics[width=0.9\textwidth]{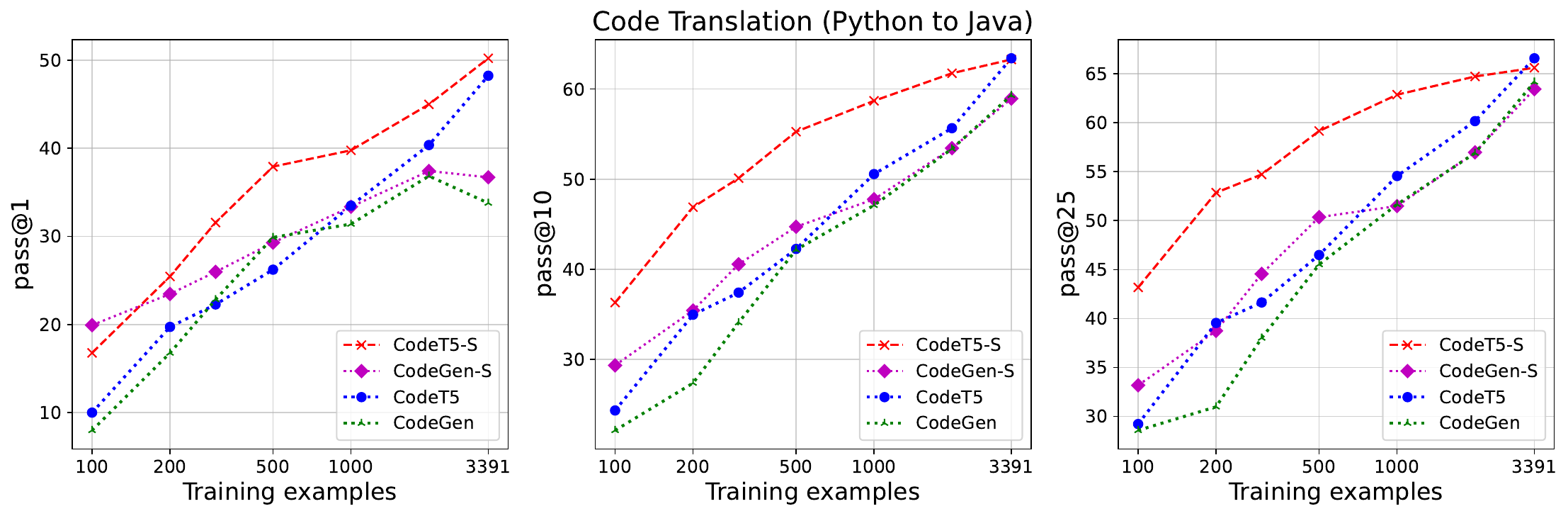}
  \caption{Code translation performance on the CodeXGLUE and TransCoder benchmarks.}
  \label{fig:codetrans-alldatasets}
\end{figure*}

In Figure \ref{fig:codetrans-alldatasets}, we show the code translation results on all benchmarks and translation directions: CodeXGLUE (Java $\leftrightarrow$ C\#) and TransCoder (Java $\leftrightarrow$ Python), under various evaluation metrics.
Similar to the findings discussed in the main text, our structured models outperform the base counterparts in most of the cases.
This observation sustains all metrics and all values of $k$ in pass@$k$.
Moreover, if a dataset does not come with test cases, the improvement demonstrated by the structured models is more significant on the fuzzy metrics (BLEU and CodeBLEU) than on the exact match.

\section{Code Generation Results}
\label{apdx:code-generation-alldatasets}

\begin{figure*}[ht]
  \centering
  \includegraphics[width=0.9\textwidth]{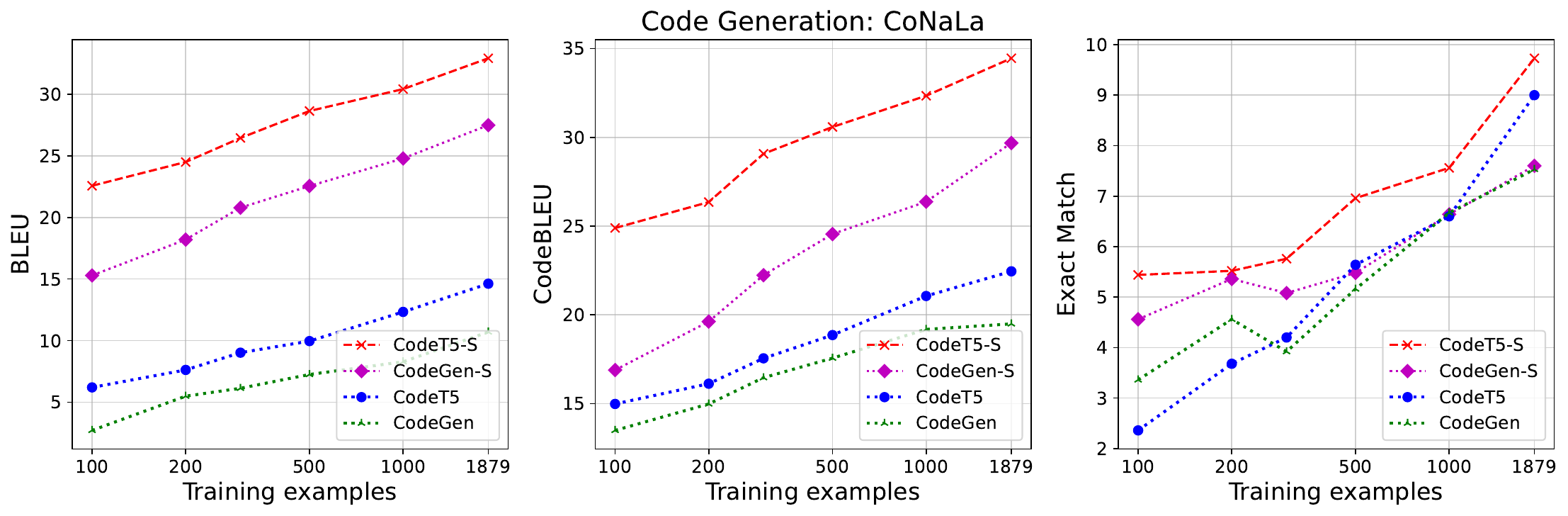} 
  \\ \vspace{5pt}
  \includegraphics[width=0.9\textwidth]{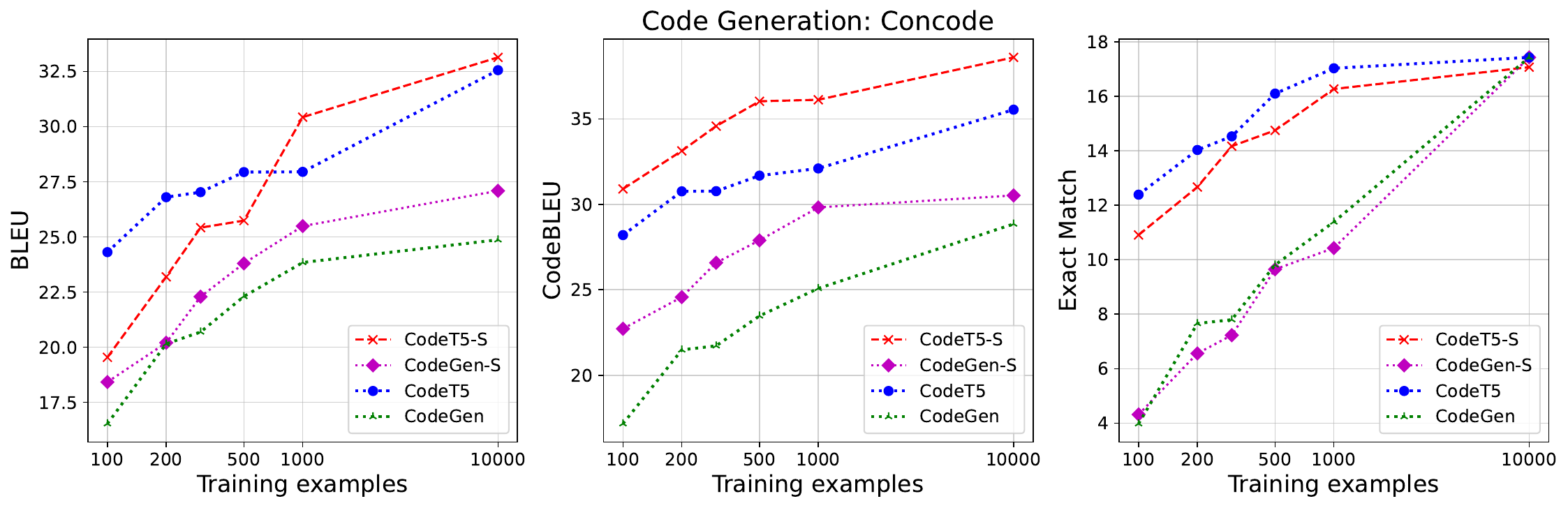}
  \\ \vspace{5pt}
  \includegraphics[width=0.9\textwidth]{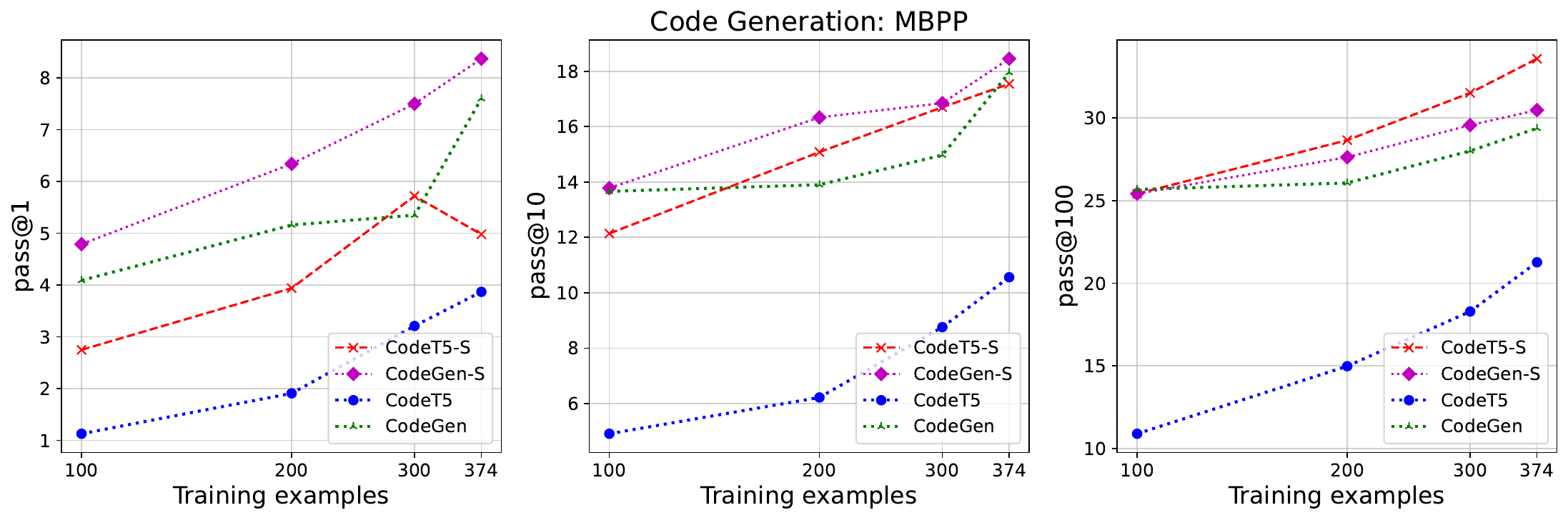} 
  \caption{Code generation performance on the CoNaLa, Concode, and MBPP benchmarks.}
  \label{fig:text2code-alldatasets}
\end{figure*}

In Figure \ref{fig:text2code-alldatasets}, we show the code generation results on all three benchmarks: CoNaLa, Concode, and MBPP, under various evaluation metrics.
Similar to the findings in the translation task, our structured models outperform the base counterparts in most of the cases.
This observation sustains all metrics and all values of $k$ in pass@$k$.

\section{Code Summarization Results}
\label{apdx:code-summarization-alllangs}

\begin{figure*}[ht]
  \centering
  \includegraphics[width=0.9\textwidth]{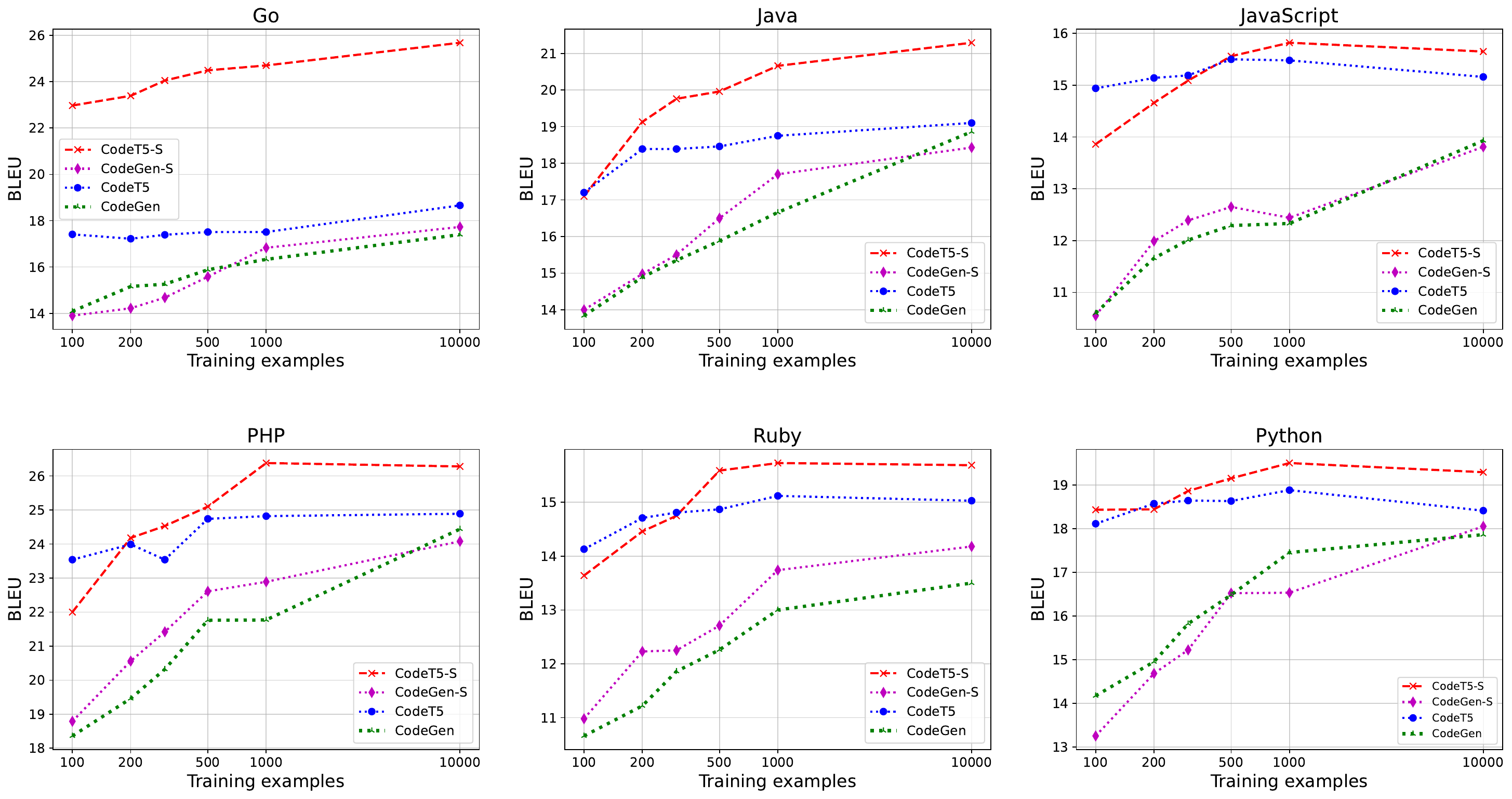}
  \caption{Code summarization performance on the CodeSearchNet benchmark.}
  \label{fig:code2text-alllangs}
\end{figure*}

In Figure \ref{fig:code2text-alllangs}, we show the code summarization results on all six languages in the CodeSearchNet benchmark.
For the language go, our structured model carries the outstanding performance improvement over the base CodeT5 model.
For other languages, the structured model also generally outperforms the base CodeT5, except for a few cases.

\input{code_examples/code-ex}

\input{code_examples/mbpp_example_1}
\input{code_examples/mbpp_example_2}

\input{code_examples/avatar_example_1}
\input{code_examples/avatar_example_2}

%% file: code_examples/code-ex.tex
\begin{figure*}
\begin{tcbitemize}[raster columns=2, raster equal height=rows, left=0.5em, right=0.5em]
\tcbitem[title=Python example]
\begin{lstlisting}[language=Python]
(*@\codehl{\#\#\# Code}@*)
def add_nums(a, b):
  c = a + b
  return c
\end{lstlisting}
\begin{lstlisting}
(*@\codehl{\#\#\# Serialized CST (pretty-printed format)}@*)
(_.module 
  (_.function_definition 
    def 
    (_.identifier add_nums identifier._) 
    (_.parameters 
      ( 
      (_.identifier a identifier._) 
      , 
      (_.identifier b identifier._) 
      ) 
    parameters._) 
    : indent 
    (_.block 
      (_.expression_statement 
        (_.assignment 
          (_.identifier c identifier._) 
          = 
          (_.binary_operator 
            (_.identifier a identifier._) 
            + 
            (_.identifier b identifier._) 
          binary_operator._) 
        assignment._) 
      expression_statement._) 
      newline 
      (_.return_statement 
        return 
        (_.identifier c identifier._) 
      return_statement._) 
      newline 
      dedent 
    block._) 
  function_definition._) 
module._)
\end{lstlisting}

\tcbitem[title=Go example]
\begin{lstlisting}[language=Go]
(*@\codehl{/// Code}@*)
func main() {
  fmt.Println("7.0/3.0 =", 7.0/3.0)
  fmt.Println(true && false)
}         
\end{lstlisting}
\begin{lstlisting}
(*@\codehl{/// Serialized CST (pretty-printed format)}@*)
(_.function_declaration 
  func 
  (_.identifier main identifier._) 
  (_.parameter_list ( ) parameter_list._) 
  (_.block 
    { 
    (_.call_expression 
      (_.selector_expression 
        (_.identifier fmt identifier._) 
        . 
        (_.field_identifier Println field_identifier._) 
      selector_expression._) 
      (_.argument_list 
        ( 
        (_.interpreted_string_literal "7.0/3.0_=" interpreted_string_literal._) 
        , 
        (_.binary_expression 
          (_.float_literal 7.0 float_literal._) 
          / 
          (_.float_literal 3.0 float_literal._) 
        binary_expression._) 
        ) 
      argument_list._) 
    call_expression._) 
    \n\n 
    (_.call_expression 
      (_.selector_expression 
        (_.identifier fmt identifier._) 
        . 
        (_.field_identifier Println field_identifier._) 
      selector_expression._) 
      (_.argument_list 
        ( 
        (_.binary_expression true && false binary_expression._) 
        ) 
      argument_list._) 
    call_expression._) 
    \n } 
  block._) 
function_declaration._)
\end{lstlisting}
\end{tcbitemize}
\caption{Example programs and the corresponding serialized CSTs.}
\label{fig:code-ex}
\end{figure*}

%% file: code_examples/mbpp_example_1.tex
\begin{figure*}[ht]
\begin{tcolorbox}[toprule at break=0pt, bottomrule at break=0pt, left=0.4em, right=0.4em, top=0em, bottom=0.2em]
\begin{lstlisting}[basicstyle=\fontsize{9}{10}\ttfamily]
(*@\codehl{Write a function to sort a list of tuples using lambda.}@*)
\end{lstlisting}
\begin{tcbitemize}[raster columns=2, raster equal height=rows, raster force size=false, left=0.4em, right=0.4em, top=0em, bottom=0em]
\tcbitem[title=CodeT5]
\begin{lstlisting}[language=Python]
def sort_list(test_list):
  (*@\textcolor{red}{res = sorted(test\_list)}@*)
  return (res)
\end{lstlisting}
\tcbitem[title=CodeGen]
\begin{lstlisting}[language=Python]
def sort_tuples(l):
  (*@\textcolor{red}{sort\_list = list(reversed(l))}@*)
  return sort_list
\end{lstlisting}
\tcbitem[title=CodeT5-Structured]
\begin{lstlisting}[language=Python]
def sort_list(test_list):
  res = list(sorted(test_list, \
                    key=lambda tup: tup[1]))
  return res
\end{lstlisting}
\tcbitem[title=CodeGen-Structured]
\begin{lstlisting}[language=Python]
def sort_list(l):
  l.sort(key=(lambda x: x[1]))
  return l
\end{lstlisting}
\end{tcbitemize}
\end{tcolorbox}
\caption{Model generations for a test sample in the MBPP dataset. Incorrectly generated parts are highlighted in red.}
\label{fig:mbpp_example_1}
\end{figure*}

%% file: code_examples/mbpp_example_2.tex
\begin{figure*}[ht]
\begin{tcolorbox}[toprule at break=0pt, bottomrule at break=0pt, left=0.4em, right=0.4em, top=0em, bottom=0.2em]
\begin{lstlisting}[basicstyle=\fontsize{9}{10}\ttfamily]
(*@\codehl{Write a function to remove all the tuples with length k.}@*)
\end{lstlisting}
\begin{tcbitemize}[raster columns=2, raster equal height=rows, raster force size=false, left=0.4em, right=0.4em, top=0em, bottom=0em]
\tcbitem[title=CodeT5]
\begin{lstlisting}[language=Python]
def remove_length(test_list, K):
  temp = [ele for ele in test_list]
  (*@\textcolor{red}{temp.sort()}@*)
  res = [ele for ele in temp if len(ele) != K]
  return (res)
\end{lstlisting}
\tcbitem[title=CodeGen]
\begin{lstlisting}[language=Python]
def remove_length(test_list, K):
  res = []
  for sub in test_list:
    if len(sub) != K:
      res.append(sub)
  res = list((*@\textcolor{red}{set(res)}@*))
  res = list(filter(lambda x: x != [], res))
  res = list(filter(lambda x: x, res))
  return (res)
\end{lstlisting}
\tcbitem[title=CodeT5-Structured]
\begin{lstlisting}[language=Python]
def remove_length(test_list, K):
  res = [ele for ele in test_list \
         if len(ele) != K]
  res = list(filter(lambda ele: \
                    len(ele) != K, res))
  return (res)
\end{lstlisting}
\tcbitem[title=CodeGen-Structured]
\begin{lstlisting}[language=Python]
def remove_length_k(nums, k):
  result = list(filter(lambda x: \
                       len(x) != k, nums))
  return (result)
\end{lstlisting}
\end{tcbitemize}
\end{tcolorbox}
\caption{Model generations for a test sample in the MBPP dataset. Incorrectly generated parts are highlighted in red.}
\label{fig:mbpp_example_2}
\end{figure*}

%% file: code_examples/avatar_example_1.tex
\begin{figure*}[ht]
\begin{tcolorbox}[toprule at break=0pt, bottomrule at break=0pt, left=0.4em, right=0.4em, top=0em, bottom=0.2em]

\begin{tcbitemize}[raster columns=1, raster equal height=rows, raster force size=false, left=0.4em, right=0.4em, top=0em, bottom=0em]
\tcbitem[title=Input Python Code]
\begin{lstlisting}[language=python]
def minInsertionStepToSortArray ( arr , N ) :
    lis = [ 0 ] * N
    for i in range ( N ) :
        lis [ i ] = 1
    for i in range ( 1 , N ) :
        for j in range ( i ) :
            if ( arr [ i ] >= arr [ j ] and lis [ i ] < lis [ j ] + 1 ) :
                lis [ i ] = lis [ j ] + 1
    max = 0
    for i in range ( N ) :
        if ( max < lis [ i ] ) :
            max = lis [ i ]
    return ( N - max )
\end{lstlisting}
\end{tcbitemize}
\begin{tcbitemize}[raster columns=2, raster equal height=rows, raster force size=false, left=0.4em, right=0.4em, top=0em, bottom=0em]
\tcbitem[title=CodeT5]
\begin{lstlisting}[language=Java]
static int minInsertionStepToSortArray(int arr[], int N) {
    int lis[] = new int[N];
    for ((*@\textcolor{red}{int i = 1}@*); i < N; i++) {
        lis[i] = 1;
    }
    for (int i = 1; i < N; i++) {
        for ((*@\textcolor{red}{int j = i; j <= i;}@*) j++)
            (*@\textcolor{red}{if (arr[i] >= arr[j])}@*) {
                (*@\textcolor{red}{lis[j]}@*) = lis[j] + 1;
            }
    }
    int max = 0;
    for ((*@\textcolor{red}{int i = 1;}@*) i < N; i++) {
        if (max < lis[i]) max = lis[i];
    }
}
return (N - max);
}
\end{lstlisting}
\tcbitem[title=CodeGen]
\begin{lstlisting}[language=Java]
static int minInsertionStepToSortArray(int[] arr, int N) {
    (*@\textcolor{red}{int lb = 0, ub = N - 1;}@*)
    for (int i = 0; (*@\textcolor{red}{i <= N}@*); i++) {
        (*@\textcolor{red}{if (arr[i] >= arr[lb] \&\& arr[i] < arr[ub] + 1) \{}@*)
            (*@\textcolor{red}{ub = i;}@*)
        (*@\textcolor{red}{\} else \{}@*)
            (*@\textcolor{red}{lb = i;}@*)
        (*@\textcolor{red}{\}}@*)
    }
    int max = 0;
    for ((*@\textcolor{red}{int i = lb; i <= N}@*); i++) {
        if (max < arr[i]) {
            max = arr[i];
        }
    }
    return (N - max);
}
\end{lstlisting}
\tcbitem[title=CodeT5-Structured]
\begin{lstlisting}[language=Java]
static int minInsertionStepToSortArray(int arr[], int N) {
    int[] lis = new int[N];
    for (int i = 0; i < N; i++) {
        lis[i] = 1;
    }
    for (int i = 1; i < N; i++) {
        for (int j = 0; j < i; j++) {
            if ((arr[i] >= arr[j] && lis[i] < lis[j] + 1)) lis[i] = lis[j] + 1;
        }
    }
    int max = 0;
    for (int i = 0; i < N; i++) {
        if (max < lis[i]) max = lis[i];
    }
    return (N - max);
}
\end{lstlisting}
\tcbitem[title=CodeGen-Structured]
\begin{lstlisting}[language=Java]
static int minInsertionStepToSortArray(int arr[], int N) {
    int[] lis = new int[N];
    for (int i = 0; i < N; i++) lis[i] = 1;
    for (int i = 1; i < N; i++) {
        for (int j = 0; j < i; j++)
            if (arr[i] >= arr[j] && lis[i] < lis[j] + 1) lis[i] = lis[j] + 1;
    }
    int max = 0;
    for (int i = 0; i < N; i++)
        if (max < lis[i]) max = lis[i];
    return (N - max);
}
\end{lstlisting}
\end{tcbitemize}
\end{tcolorbox}
\caption{Model translations for a test sample in the TransCoder (Python $\rightarrow$ Java) dataset with Cyclomatic Complexity = 8 for the input Python program. Incorrectly generated parts are highlighted in red.}
\label{fig:avatar_example_1}
\end{figure*}

%% file: code_examples/avatar_example_2.tex
\begin{figure*}[ht]
\begin{tcolorbox}[toprule at break=0pt, bottomrule at break=0pt, left=0.4em, right=0.4em, top=0em, bottom=0.2em]

\begin{tcbitemize}[raster columns=1, raster equal height=rows, raster force size=false, left=0.4em, right=0.4em, top=0em, bottom=0em]
\tcbitem[title=Input Python Code]
\begin{lstlisting}[language=python]
def findSmallestDifference ( A , B , m , n ) :
    A.sort ( )
    B.sort ( )
    a = 0
    b = 0
    result = sys.maxsize
    while ( a < m and b < n ) :
        if ( abs ( A [ a ] - B [ b ] ) < result ) :
            result = abs ( A [ a ] - B [ b ] )
        if ( A [ a ] < B [ b ] ) :
            a += 1
        else :
            b += 1
    return result
\end{lstlisting}
\end{tcbitemize}
\begin{tcbitemize}[raster columns=2, raster equal height=rows, raster force size=false, left=0.4em, right=0.4em, top=0em, bottom=0em]
\tcbitem[title=CodeT5]
\begin{lstlisting}[language=Java]
static int findSmallestDifference(int A[], int B[], int m, int n) {
    Arrays.sort(A);
    Arrays.sort(B);
    int result = (*@\textcolor{red}{Math.abs(A[a] - B[b]);}@*)
    while (a < m && b < n) {
        if (abs(A[a] - B[b]) < result) {
            result = Math.abs(A[a] - B[b]);
            (*@\textcolor{red}{a += 1;}@*)
        } else {
            b += 1;
        }
    }
    return result;
}
\end{lstlisting}
\tcbitem[title=CodeGen]
\begin{lstlisting}[language=Java]
static long findSmallestDifference(long A[], long B[], int m, int n) {
    Arrays.sort(A);
    Arrays.sort(B);
    long result = 0;
    int a = 0, b = 0;
    (*@\textcolor{red}{int min\_a = 0, min\_b = 0;}@*)
    (*@\textcolor{red}{int count = 0;  int min\_count = 0;  long sum = 0;}@*)
    for (; a < m && b < n; (*@\textcolor{red}{a++, b++}@*)) {
        (*@\textcolor{red}{if (A[a] < B[b])}@*) {
            (*@\textcolor{red}{result = A[a]; min\_a = a;}@*)
            (*@\textcolor{red}{min\_b = b; count++;}@*)
            (*@\textcolor{red}{min\_count = count; sum += A[a];}@*)
        } else {
            (*@\textcolor{red}{result = B[b]; min\_a = b;}@*)
            (*@\textcolor{red}{min\_b = a; count++;}@*)
            (*@\textcolor{red}{min\_count = count; sum += B[b];}@*)
        }
    }
    return result;
}
\end{lstlisting}
\tcbitem[title=CodeT5-Structured]
\begin{lstlisting}[language=Java]
static int findSmallestDifference(int A[], int B[], int m, int n) {
    Arrays.sort(A);
    Arrays.sort(B);
    int a = 0, b = 0;
    int result = Integer.MAX_VALUE;
    while (a < m && b < n) {
        if (Math.abs(A[a] - B[b]) < result) {
            result = Math.abs(A[a] - B[b]);
        }
        if (A[a] < B[b]) {
            a++;
        } else {
            b++;
        }
    }
    return result;
}
\end{lstlisting}
\tcbitem[title=CodeGen-Structured]
\begin{lstlisting}[language=Java]
static int findSmallestDifference(int A[], int B[], int m, int n) {
    Arrays.sort(A);
    Arrays.sort(B);
    int a = 0, b = 0;
    int result = Integer.MAX_VALUE;
    while (a < m && b < n) {
        if (Math.abs(A[a] - B[b]) < result) result = Math.abs(A[a] - B[b]);
        if (A[a] < B[b]) a++;
        else b++;
    }
    return result;
}
\end{lstlisting}
\end{tcbitemize}
\end{tcolorbox}
\caption{Model translations for a test sample in the TransCoder (Python $\rightarrow$ Java) dataset with Cyclomatic Complexity = 5 for the input Python program. Incorrectly generated parts are highlighted in red.}
\label{fig:avatar_example_2}
\end{figure*}